\documentclass[letterpaper]{article} 
\usepackage{aaai2026}  
\usepackage{times}  
\usepackage{helvet}  
\usepackage{courier}  
\usepackage[hyphens]{url}  
\usepackage{graphicx} 
\usepackage{placeins} 
\usepackage{natbib}  
\usepackage{caption} 
\usepackage{amsmath}
\usepackage{dsfont}
\usepackage{booktabs}  
\usepackage{tabularx}  
\usepackage{caption}   
\usepackage{tabularx}  
\usepackage{caption}   
\usepackage{multirow}  

\usepackage{enumitem}

\usepackage[ruled,linesnumbered]{algorithm2e}
\usepackage{arydshln}
\usepackage{makecell}

\usepackage{newfloat}
\usepackage{listings}
\DeclareCaptionStyle{ruled}{labelfont=normalfont,labelsep=colon,strut=off} 
\title{ToolSample: Dual Dynamic Sampling Methods with Curriculum Learning for RL-based Tool Learning}

\author{%
Zihao Feng$^{1,2}$\thanks{Equal contribution}\thanks{Zihao Feng was an intern at Tencent during the preparation of this work}, Xiaoxue Wang$^{2*}$, Bowen Wu$^{2}$, Hailong Cao$^{1}$,\\Tiejun Zhao$^{1}$\thanks{Corresponding author}, Qun Yu$^{2}$, Baoxun Wang$^{2}$} 

\affiliations{
    \textsuperscript{\rm 1}Faculty of Computing, Harbin Institute of Technology\\
    \textsuperscript{\rm 2}Platform and Content Group, Tencent\\
    fengzihaogl@outlook.com \\
    \{yukixxwang, jasonbwwu, sparkyu, asulewang\}@tencent.com \\
    \{caohailong, tjzhao\}@hit.edu.cn

}

\usepackage{bibentry}

\begin{document}

\maketitle

\begin{abstract}
While reinforcement learning (RL) is increasingly used for LLM-based tool learning, its efficiency is often hampered by an overabundance of simple samples that provide diminishing learning value as training progresses. Existing dynamic sampling techniques are ill-suited for the multi-task structure and fine-grained reward mechanisms inherent to tool learning. This paper introduces Dynamic Sampling with Curriculum Learning (DSCL), a framework specifically designed to address this challenge by targeting the unique characteristics of tool learning: its multiple interdependent sub-tasks and multi-valued reward functions. DSCL features two core components: Reward-Based Dynamic Sampling, which uses multi-dimensional reward statistics (mean and variance) to prioritize valuable data, and Task-Based Dynamic Curriculum Learning, which adaptively focuses training on less-mastered sub-tasks. Through extensive experiments, we demonstrate that DSCL significantly improves training efficiency and model performance over strong baselines, achieving a 3.29\% improvement on the BFCLv3 benchmark. Our method provides a tailored solution that effectively leverages the complex reward signals and sub-task dynamics within tool learning to achieve superior results.
\end{abstract}

\section{Introduction}
Large Language Models (LLMs) have shown promising capabilities in complex tasks, including tool learning~\cite{qin2024tool,kumar2025llm,qu2025tool}. Tool learning aims to unleash the power of LLMs by enabling them to interact with external APIs to complete complex human instructions~\cite{tang2023toolalpaca, zeng2025toolace}. 
Meanwhile, reinforcement learning (RL) has emerged as a key strategy for enhancing the instruction-following and reasoning capabilities of LLMs, as demonstrated by the performance of models such as DeepSeek-R1~\cite{guo2025deepseek}. This success has validated the viability of using RL to train agents for complex, multistep tasks, thereby accelerating their adoption for the specific challenge of tool learning~\cite{feng2025retool,feng2025improving,chen2025empirical,li2025torl}.
  
\begin{figure}[t] 
\centering
\includegraphics[width=0.5\textwidth]{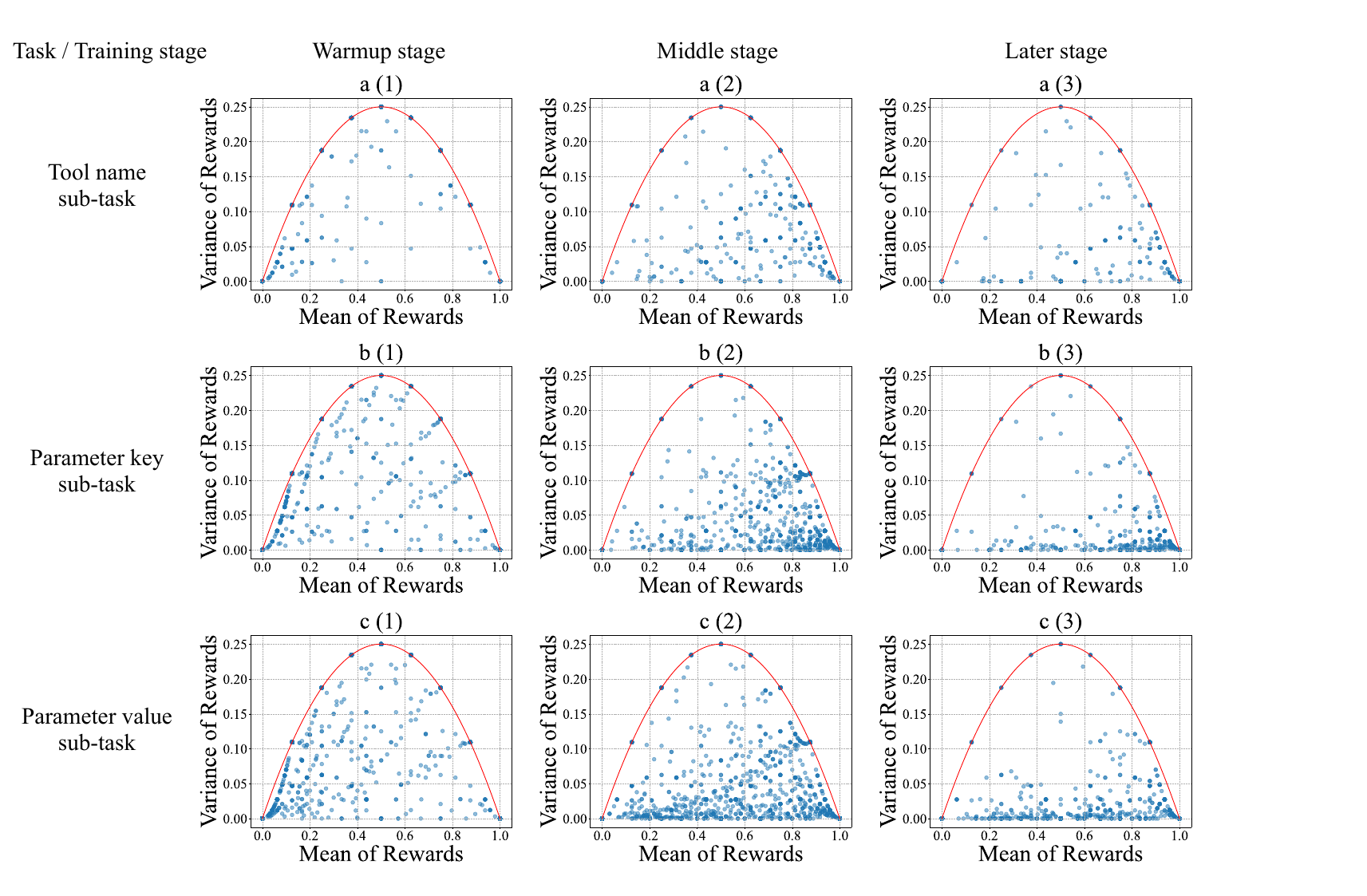} 
\caption{
Mean-Variance Scatter Plots of training rewards across different stages of ToolRL training. (Warmup: The stage where the model's format prediction is almost correct; Middle and Later: The middle and final stages of training, respectively)
Key observations include: 1) In the Middle and Later stages, the reward variance for many samples collapses towards zero;
2) Unlike the distribution (red) assuming a binary reward, where variance is a direct function of the mean, the actual mean-variance distribution (blue) of tool-learning exhibits a wide range of variances for any given mean. 
3) Different sub-tasks (e.g., a(2) and c(2)) exhibit asynchronous convergence within the same stage. 
} 
\label{fig:introduction}
\end{figure}

To better adapt RL for tool learning task, \citet{qian2025toolrl} have proposed ToolRL method that incorporates independent reward design for tool selection, tool parameter key extraction, and parameter value completion to provide fine-grained feedback to the model. 
While effective, this approach is limited to improvements in the aspect of the reward, lacking any sample-level filtering to optimize the RL training process.
Figure~\ref{fig:introduction} illustrates the distribution of sample reward means and variances at different stages of ToolRL training. It shows that during the middle to late stages of training, the reward variance for many samples approaches zero, particularly for parameter key and value prediction (subfigure b(2-3) and c(2-3)). Previous research has demonstrated that such homogeneous samples provide limited benefit to training effectiveness, and when their proportion becomes excessive, they may even impede further model improvement~\cite{team2025kimi,bae2025online}.

To address this issue, sample selection methods that prioritize valuable data have proven effective in various RL tasks such as mathematics \cite{yu2025dapo} and code generation \cite{liu2025prorl}, but their application in tool learning remains unexplored in current public research.
The tool learning task differs fundamentally from those tasks in two critical aspects: 1)it comprises multiple interdependent sub-tasks 2) it benefits more from fine-grained than coarse-grained reward mechanisms \cite{qian2025toolrl}, which make existing dynamic sampling methods less suitable for tool learning task.
This transition from a binary to a multi-valued reward fundamentally alters the relationship between the mean and variance. As Figure~\ref{fig:introduction} shows, a binary reward mathematically locks the variance to the mean (red curves), making a single metric (either mean or variance) sufficient for evaluating data samples. In contrast, ToolRL's multi-valued reward setting decouples them (blue scatter plots), allowing a given mean to correspond to a wide range of variances. This enables the mean and variance to provide independent signals from distinct dimensions to guide data sampling during training.
Moreover, Figure~\ref{fig:introduction} also illustrates that, sub-tasks  (e.g., subfigures a(2) and c(2)) exhibit asynchronous convergence as shown in Figure~\ref{fig:introduction}.
Therefore, an effective dynamic sampling method designed for tool learning must:
(1) fully consider multi-dimensional data information;
(2) dynamically adjust sampling strategy based on the current training stage and the degree of convergence of different sub-tasks; and (3) balance the sample distribution across sub-tasks to avoid over-focusing on well-learned components at the expense of those that still require improvement.

To fully exploit the potential of fine-grained reward mechanisms within tool learning, we propose customized dynamic sampling methods with curriculum learning (DSCL). This methodology is composed of two strategies: Reward-Based Dynamic Sampling (RDS) and Task-Based Dynamic Curriculum Learning (TDCL).
Within RDS, we propose comprehensive metrics to capture the task's reward diversity. These dynamically track three dimensions: the reward mean and variance of a group of rollouts, and the mean reward variance across the entire training. This multi-dimensional approach facilitates an assessment of not only the instantaneous difficulty and stability of the samples but also their evolutionary trajectories throughout the learning history, enabling more efficient and exploratory training. 
Additionally, the TDCL method emphasizes targeted sub-tasks according to the training status of samples, so as to help model training via providing diverse variances between rollouts.
Overall, this paper offers the following contributions:

\begin{itemize}
    \item We introduce Dynamic Sampling with Curriculum Learning (DSCL), the first dynamic sampling method specifically tailored to the unique characteristics of tool learning.
    \item DSCL consists of two novel sub-methods: RDS, which dynamically samples training data according to multiple dimensions of the mean and variance of rewards, and TDCL, which constructs a three-stage curriculum by exploiting sub-task dependencies.
    \item We conduct comprehensive experiments to evaluate our DSCL method against several strong baselines. The results, supported by our in-depth analysis, confirm its significant effectiveness and superiority.
\end{itemize}

\section{Related Work}
Our research builds upon recent advancements in two key areas: the development of tool learning in LLMs and the optimization of the training process in RL.

\subsection{Tool Learning}

The ability of LLMs to interact with external tools has evolved significantly. Foundational work like ToolLLM \cite{qin2023toolllm} pioneered the field by creating large-scale benchmarks for supervised fine-tuning. To overcome the limitations of static imitation, subsequent research has employed Reinforcement Learning to optimize sequential tool use, as seen in StepTool \cite{yu2024steptool}, Search-R1 \cite{jin2025search}, and Tool-N1~\cite{zhang2025nemotron}, etc. The success of RL, however, is highly dependent on effective reward engineering; previous works such as ToolRL~\cite{qian2025toolrl} and ARTIST~\cite{singh2025agentic} focus on systematically addressing this challenge.

\subsection{Sample Selection for RL}
To improve the efficiency of RL-based training, curriculum learning and dynamic sampling strategies are two widely used strategies. A core idea of ~\textbf{Curriculum Learning} is to structure training from simple to complex tasks, as demonstrated by Confucius \cite{gao2024confucius}. A prominent strategy within this is difficulty-aware sampling, which dynamically focuses on problems of appropriate difficulty. Its effectiveness was shown at scale by Kimi K1.5 \cite{team2025kimi}, with focused studies like RCS \cite{feng2025improving} and online difficulty filtering \cite{bae2025online} confirming that an intermediate difficulty level maximizes learning efficiency. 
~\textbf{Dynamic sampling} organizes samples from an alternative perspective. Approaches like POLARIS \cite{Polaris2025} employ adaptive filtering to enhance exploration, while frameworks such as DAPO \cite{yu2025dapo} and VAPO \cite{yue2025vapo} integrate dynamic sampling directly into the policy optimization process to address scaling challenges. \citet{razin2025makes} have also theoretically explored the detrimental effects of low-variance rewards on training effectiveness.

\begin{figure*}[htbp]  
\centering
\includegraphics[width=1\textwidth]{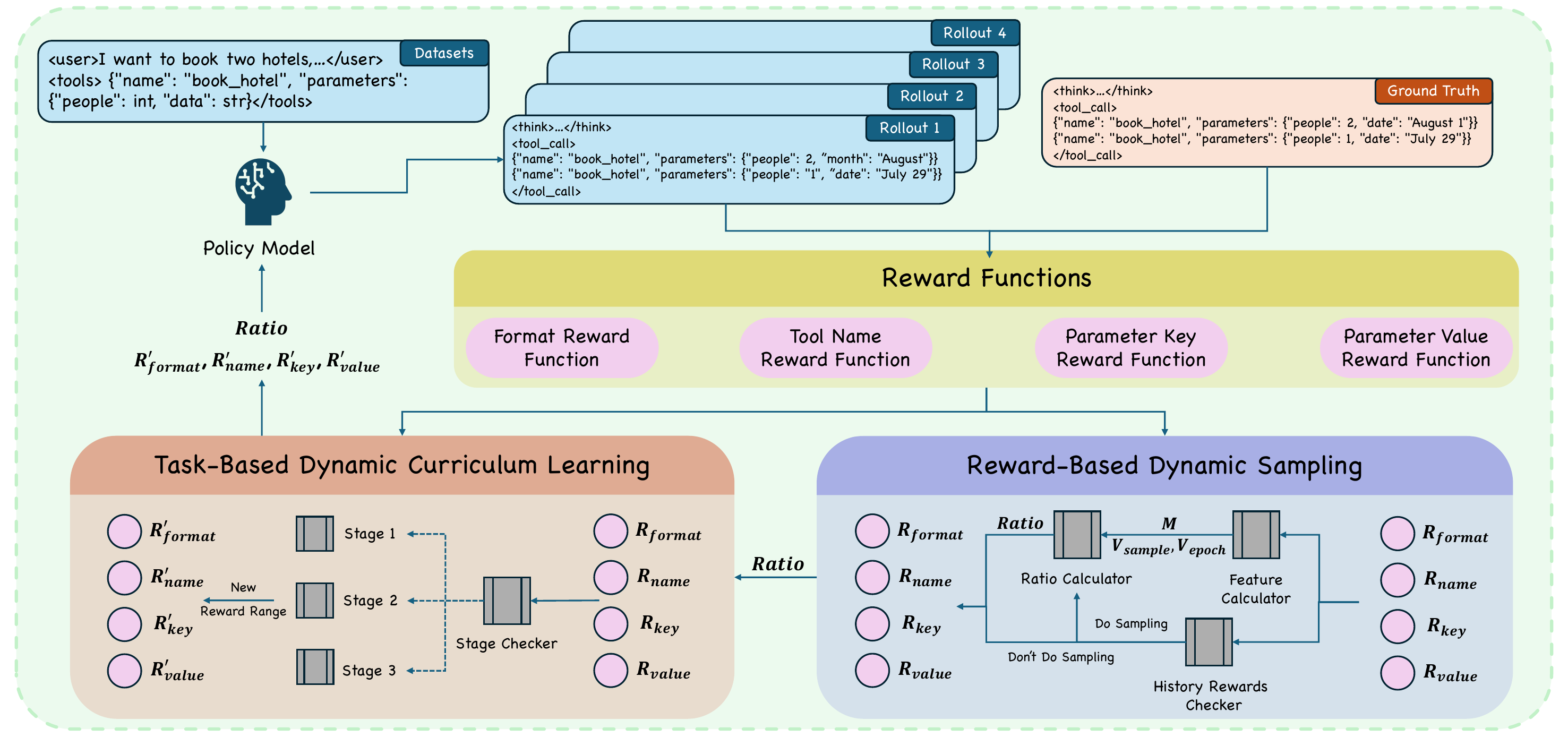} 
\caption{The training framework of our proposed method.} 
\label{framework}
\end{figure*}

\section{Reward Design for Tool Learning}
\citet{qian2025toolrl} have done a comprehensive study on reward design. They have divided the tool learning task into four sub-tasks for GRPO training. We maintain their definitions of rewards and modify each sub-task into an independent representation. For a specific query, the reward functions compare the predicted tool calls $\widehat{Y}=\{\hat{y}_{1},...,\hat{y}_{m}\}$ with the ground-truth tools $Y=\{y_{1},...,y_{m}\}$. It includes the following four components:

\paragraph{Format Reward} The format reward $R_{format} \in \{0,1\}$ checks whether the model's response meets the requirement for both content and order:

\begin{equation}
R_{format} =
\left\{  
\begin{array}{ll}  
1, & \text{if the format meet all requirements} \\  
0, & \text{otherwise} 
\end{array}.
\right. 
\end{equation}

\paragraph{Tool Name Reward} The tool name reward $R_{name}$ evaluates the correctness of each tool's name. The $N_{\widehat{Y}}$ and $N_{Y}$ represent the predicted tool names and the ground truth, respectively.

\begin{equation}
    R_{name} = \frac{|N_{\widehat{Y}} \cap N_{Y}|}{|N_{\widehat{Y}} \cup N_{Y}|} \in [0,1].
\end{equation}

\paragraph{Parameter Key Reward}
The parameter key reward $R_{key}$ checks whether the key of the parameter of each tool is correct, which is on the premise that the tool name is correct:


\begin{equation}
    R_{key} = \sum_{y_{i}\in Y}\frac{|K(f(y_{i})) \cap K(y_{i})|}{|K(y_{i})|+|K(f(y_{i}))-K(y_{i})|} \in [0,|Y|],
\end{equation}



\noindent where function $f(y_{i})$ selects the best matches tool $\hat{y}_{i}$ with $y_{i}$ in the predicted tools. If there is no match, it returns empty. The $K$ represents the set of keys of a specific tool.

\paragraph{Parameter Value Reward}
The parameter value reward $R_{value} \in [0,\sum_{y_{i}\in Y}|K(y_{i})|]$ evaluates the correctness of parameter values for all the matching keys. 


\begin{equation}
\begin{split}
    R_{value} =\sum_{y_{i}\in Y} \sum_{k\in K(y_{i})}\frac{ \mathds{1} [V(f(y_{i})^{k}) = V(y_{i}^{k})]}{|V(y_{i})|+|V(f(y_{i}))-V(y_{i})|},
\end{split}
\end{equation}

\noindent where $V(y_{j}^{k})$ represents the value that is corresponding to the tool $y_{j}$'s key $k$.

\paragraph{Total Reward} The final reward first maps the sum of the rewards of the last three sub-tasks to the interval $[-3,4]$, and then adds it with $R_{format}$ to get the final reward $R$.

\begin{equation}
\label{eqr}
    R = R_{format} + \mathcal{M}_{-3}^{3}(R_{name} + R_{key} + R_{value}),
\end{equation}


\begin{equation}
\label{eqr}
    \mathcal{M}_{min}^{max}(R)=\left \{\frac{max-min}{R_{max}-R_{min}}*(r_{i}-R_{min})+min \right \},
\end{equation}

\noindent where $\mathcal{M}$ represents the mapping function, $r_{i}$ represents an element in the set $R$, $R_{max}$ and $R_{min}$ represent the upper and lower bounds of the set $R$.


\section{Methodology}
In this section, we will introduce our proposed Dynamic Sampling with Curriculum Learning (DSCL) for RL-based tool learning. The DSCL method contains two strategies, Reward-Based Dynamic Sampling (\textbf{RDS}) and Task-Based Dynamic Curriculum Learning (\textbf{TDCL}). We will introduce these two methods separately and then explain how they constitute the DSCL method.

\subsection{Reward-Based Dynamic Sampling (RDS)}
We will first describe the basic formulation of the tool learning task. Let $D=\{d_{1},...,d_{k}\}$ represents the dataset for training. For each data $d_{i}$ at the $j$-$th$ epoch, we generate $G$ samples and calculate their rewards $\mathcal{R}^{i,j}=\{R^{i,j}_{1},...,R^{i,j}_{G}\}$.



The RDS method is initiated only after the model achieves stable proficiency in tool prediction\footnote{We operationalize this stability criterion as the mean reward of seven consecutive batches surpassing a threshold of 1.0 in practice.}, a strategic delay designed to prevent the premature rejection of a large volume of samples. Subsequently, it quantifies sample importance by integrating a sample's current training status with the potential learning advantage it offers, thereby enabling dynamic sampling across the entire training corpus.
To achieve this, for each data $d_{i}$ at the $j$-$th$ epoch, we select the following three indicators to capture the training state and reward changes, so as to better dynamically sample the data:

\paragraph{Mean Reward of Samples} Mean reward for all rollouts:

\begin{equation}
    M^{i,j} = Mean(\{R^{i,j}_{1},...,R^{i,j}_{G}\}).
\end{equation}

\paragraph{Variance of Samples} The sample-level variance $V_{sample}$ for each data at each epoch:

\begin{equation}
    V_{sample}^{i,j} = Var(\{R^{i,j}_{1},...,R^{i,j}_{G}\}).
\end{equation}

\paragraph{Variance of Epoches} The epoch-level variance, i.e., the variance of all $M^{i,j}$ during past epochs:

\begin{equation}
    V_{epoch}^{i,j} = Var(\{M^{i,1},...,M^{i,j}\}).
\end{equation}


Depends on these three indicators, we have dynamically divided the data into the following four categories:
\begin{itemize}
    \item a. Easy data: $M^{i,j} = 4$.
    \item b. Hard data: $M^{i,j} < t_{mean}$.
    \begin{itemize}
        \item b.1 $V_{sample}^{i,j} > t_{var}$ or $V_{epoch}^{i,j} > t_{var}$.
        \item b.2 $V_{sample}^{i,j} < t_{var}$ and $V_{epoch}^{i,j} < t_{var}$.
    \end{itemize}
    \item c. Intermediate data: $M^{i,j} \in (t_{mean},4)$ .
    \begin{itemize}
        \item c.1 $V_{sample}^{i,j} > t_{var}$ and $V_{epoch}^{i,j} > t_{var}$.
        \item c.2 $V_{sample}^{i,j} < t_{var}$ or $V_{epoch}^{i,j} < t_{var}$.
    \end{itemize}
\end{itemize}

We assign different importance as the sampling ratio according to the category of each data $d_{i}$: 
\begin{equation}
Ratio_{i} =
\left\{  
\begin{array}{ll}  
0.0, & d_{i} \in \{a,b.2\}\\
0.5, & d_{i} \in \{c.2\}\\
1.0, & d_{i} \in \{b.1,c.1\}
\end{array}.
\right. 
\end{equation}

\noindent Specifically, we discard samples that are either too challenging for the current model or have already been fully mastered. For challenging samples, we fully retain them if they exhibit either effective diversity across rollouts or exploratory behavior from the model compared to past epochs. For less challenging samples, however, we apply a stricter requirement, retaining them only if they satisfy both conditions. 
Other samples are partially retained, as relying exclusively on highly informative or exploratory samples can lead to training instability~\cite{dang2025reinforcement}. Furthermore, $t_{mean}$ and $t_{var}$ are treated as hyperparameters that are adjusted during training. This allows the method to accommodate the sensitivity of the reinforcement learning process to variations in data and model states~\cite{wang2025tina}.

\subsection{Task-Based Dynamic Curriculum Learning (TDCL)}


The sub-tasks within the tool learning framework exhibit varying levels of difficulty and a natural progressive relationship. For instance, the format task is foundational, providing structured data for all subsequent operations. We further analyze the training loss and find a difficulty hierarchy: 1) Extracting the tool name and its parameter keys is usually synchronous, i.e., successfully predicting the tool name will link it to the corresponding parameter keys; 2) Accurately extracting parameter values is significantly more difficult.

Based on these observations, we design a three-stage dynamic curriculum learning strategy. At each stage, we adjust the reward weights for these sub-tasks, as defined in Eq.~\ref{eqr}, to systematically shift the training focus from simpler to more complex objectives:



\begin{itemize}
    \item \textbf{1: } We increase the $\lambda_{format}$ to 2.5 times its original value and reduce other three parameters by half and the penalty to enhance the learning of the format in the first stage.

\begin{equation}
\begin{split}
    R = & 2.5*R_{format} + \mathcal{M}_{0}^{1.5}(0.5*R_{name} \\
    &+ 0.5*R_{key} + 0.5*R_{value})
\end{split}.
\end{equation}

    \item \textbf{2: } In the second stage, we reduce the $\lambda_{format}$ by half and add the penalty, maintain the $R_{value}$ as stage 1 and increase the other two parameters to 1.5 times.

\begin{equation}
\begin{split}
    R = & \mathcal{M}_{-1}^{0.5}(R_{format}) + \mathcal{M}_{0}^{3.5}(1.5*R_{name} \\
    &+ 1.5*R_{key} + 0.5*R_{value})
\end{split}.
\end{equation}

    \item \textbf{3: } In the last stage, we maintain the $\lambda_{format}$ as stage 2, increase $\lambda_{value}$ to 2.5 times its original value and reduce other two parameters.

\begin{equation}
\begin{split}
    R = & \mathcal{M}_{-1}^{0.5}(R_{format}) + \mathcal{M}_{0}^{3.5}(0.5*R_{name} \\
    &+ 0.5*R_{key} + 2.5*R_{value})
\end{split}.
\end{equation}

\end{itemize}
 
\noindent Besides, we have recorded the rewards of each sub-task of several latest batches to determine which stage to execute, which is the same as the warmup before performing RDS.

\subsection{Combination of RDS and TDCL}
We combine the two strategies to form our DSCL method. Specifically, we first execute the RDS strategy to obtain the sampling ratios based on the original reward and then execute the TDCL strategy. 
The overall training framework is illustrated in Figure~\ref{framework}, while the detailed execution process is outlined in Algorithm~\ref{algorithm}.

\begin{algorithm}[]  
	\caption{Dynamic Sampling with Curriculum Learning (DSCL)}
	\label{algorithm}
	\LinesNumbered 
	\KwIn{Training dataset $\{D,Y\}$, GRPO sampling number $G$}
	\For{$step=1,...,step\_num$}{
            Sample a batch $D_{b}$ from $D$ \;
            $\widehat{Y}=generate\_response(D_{b},G)$ \;            $R,R_{format},R_{name},R_{key},R_{value}=calc\_reward(Y,\widehat{Y})$ \;   
            $Ratio=RDS(R,R_{format},R_{name},R_{key},R_{value})$ \;            $R,R_{format},R_{name},R_{key},R_{value}=TDCL(R,R_{format},R_{name},R_{key},R_{value})$ \;
            $\hat{A_{b}} = calc\_advantage(R)$ \;
            $\hat{A_{b}} = \hat{A_{b}} * Ratio$ \;
            $update\_model(\hat{A_{b}})$	
        }
\end{algorithm}

\section{Experiments Settings}

\subsection{Training Dataset}
For the reinforcement learning (RL) phase, we adopt the data composition and processing methodology established by the ToolRL framework \cite{qian2025toolrl}. This approach involves creating a balanced and composite dataset by sampling from three specialized sources. The rationale behind this curated blend is to expose the model to a comprehensive range of challenges, thereby fostering a robust and generalizable tool-use capability. The composition of our training data and the primary contribution of each source are detailed in Supplementary Material.

This strategically designed dataset aims to:

\begin{itemize}
    \item \textbf{ToolACE} \cite{liu2024toolace}: Build a foundational understanding of diverse and complex tool interactions.
    \item \textbf{Hammer} \cite{lin2024hammer}: Promote deep semantic reasoning over superficial name-matching.
    \item \textbf{xLAM} \cite{zhang2024xlam}: Develop advanced strategic planning for multi-step, complex tasks.
\end{itemize}

Following the ToolRL strategy, we sample 2K examples from ToolACE and 1K each from Hammer and xLAM. Besides, multi-step trajectories from these datasets are decomposed into single-step instances, with prior dialogue history preserved to maintain context.

\subsection{Evaluation Dataset}
To ensure a rigorous and directly comparable evaluation, we assess our method on the Berkeley Function Calling Leaderboard (BFCL) \cite{patil2025bfcl} and API-Bank \cite{li2023api}, the same benchmarks used in ToolRL \cite{qian2025toolrl}. We do 5 runs for each method and report the average score of the accuracies ($\%$).

\paragraph{BFCL: Analyzing Core Tool Invocation Mechanics.} BFCL V3 \cite{patil2025bfcl} is utilized to measure the fundamental correctness of tool use. Its principal strength is a diagnostic, multi-faceted structure that deconstructs a single tool use into granular dimensions, as detailed in Supplementary Material.
This approach is critical for isolating specific failure modes—such as distinguishing an error in tool selection from one in parameter formatting—thereby enabling a precise, component-level analysis of our model's fidelity.

\paragraph{API-Bank: Evaluating Multi-Step Planning and Reasoning.} Complementing BFCL's focus on mechanics, the API-Bank benchmark\cite{li2023api} evaluates higher-order reasoning within realistic, multi-turn dialogues. It employs a hierarchical evaluation that tests progressively complex skills, from basic tool execution to autonomous, multi-step planning, with a full breakdown provided in Supplementary Material.
The benchmark's foundation on manually annotated, authentic API interactions allows us to rigorously validate our model's improvements in strategic tool use.

\subsection{Baselines}
To rigorously evaluate the efficacy of our proposed method, we compare it against a set of carefully selected baselines that represent different levels of capability and distinct state-of-the-art philosophies:

\begin{itemize}
    \item \textbf{Qwen2.5-7B-Instruct} \cite{qwen2}: We use the original Qwen2.5-7B-Instruct without any additional fine-tuning or RL as the foundation model for our method and all the other baselines.
    \item \textbf{Tool-N1} \citep{zhang2025nemotron}: Tool-N1 designs binary reward function for tool learning task. They do not consider sub-tasks separately, but directly evaluate the correctness of the generated results.
    \item \textbf{ToolRL} \citep{qian2025toolrl}: ToolRL is an RL framework specifically tailored for tool-use tasks, also built on GRPO. Its primary contribution is a fine-grained reward signal that meticulously evaluates both the structural format and the semantic correctness of tool invocations.
    \item \textbf{DAPO} \citep{yu2025dapo}: An efficient RL baseline that improves training utility by using Dynamic Sampling to filter uninformative trajectories and Clip-Higher to enhance sample diversity.
    \item \textbf{Sampling by max variance (SMV)} \citep{xu2025not}: A data filtering technique that maximizes the reward variance of training batches by selecting a diverse mix of high-reward and low-reward rollouts, thus creating a more informative training signal.
    \item \textbf{Sampling by mean reward (SMR)} \citep{bae2025online}: A dynamic curriculum that uses the model's real-time pass rate (mean reward) to filter out problems that are too easy or too hard, thereby focusing training within the optimal learning zone.
\end{itemize}

Since most of the sampling methods in the baselines have not been used on tool learning task, we implement the results of these methods on tool learning task based on their open-source codes. For the last two baselines, we have conduct the method with curriculum learning based on our methodology.

\section{Experiments}

\begin{table*}[]
\centering
\begin{tabular}{lcccccc}
\toprule
Model                  & \makecell{Overal\\ Acc} & \makecell{Non-Live AST\\ Acc} & \makecell{Live\\ Acc} & \makecell{Multi Turn\\ Acc} & \makecell{Relevance\\ Detection} & \makecell{Irrelevance\\ Detection} \\\midrule
Qwen2.5-7B-Instruct    & 47.68  & 68.98    & 63.31     & 8.88  & 72.22 & 71.93          \\
Tool-N1  &  53.91 & 77.50 &	73.39 &	10.00 &	77.78  &	77.85 \\
ToolRL   & 56.96  & 85.21  & 73.39  & 13.25  & \textbf{83.33}  & 75.87    \\
\ \ \ \ +DAPO  & 47.35  & 68.27 & 64.55 & 	7.38 & 77.78 & 71.80 \\ 
\ \ \ \ +SMV  & 57.03  & 85.76   &73.79   & 12.62   &\textbf{83.33} & 77.59   \\
\ \ \ \ +SMR  & 58.18  & 86.08   &74.87   & 13.74   &\textbf{83.33} & 75.43   \\ \hdashline
\ \ \ \ +RDS & 59.55 & 85.54 & 76.10 & 17.38 & \textbf{83.33} & \textbf{79.86}    \\
\ \ \ \ +TDCL  & 59.95  & 86.25   &76.06   & 18.13   &\textbf{83.33} & 79.54   \\
\ \ \ \ +DSCL (RDS+TDCL) & \textbf{60.25}  & \textbf{86.42}   &\textbf{76.85}   & \textbf{18.50}   & \textbf{83.33} & 78.44   \\
\bottomrule
\end{tabular}
\caption{
Results on the BFCL V3 dataset. The "\textbf{bold}" indicates the best score among all the methods of each sub-task.
}
\label{maintable}
\end{table*}

\begin{table}[]
\centering
\small
\setlength{\tabcolsep}{0.8mm}
\begin{tabular}{lcccc}
\toprule
Model                  & \makecell{Overal Acc} & Level 1 & Level 2 & Level 3 \\\midrule
Qwen2.5-7B-Instruct    & 58.29  & 65.41    & 43.28     & 44.27        \\
Tool-N1  &  60.47 & 70.68 &	46.27 &	36.64  \\
ToolRL   & 61.81  & 72.18   &56.72   & 32.82     \\
\ \ \ \ +DAPO  & 58.96  & 66.17 & 41.79 & 	\textbf{45.80}  \\ 
\ \ \ \ +SMV  & 61.64  & 69.67   &49.25   & 43.51    \\
\ \ \ \ +SMR  & 62.31  & 72.43   &56.72   & 34.35    \\ \hdashline
\ \ \ \ +RDS & 63.48  & 72.18   & 56.72   & 40.46    \\
\ \ \ \ +TDCL  & 63.65  & \textbf{74.19}   &61.19   & 32.82   \\
\ \ \ \ +DSCL (RDS+TDCL) & \textbf{64.99}  & 73.18   &\textbf{65.67}   & 39.69     \\
\bottomrule
\end{tabular}
\caption{
Results on the API-Bank dataset.
}
\label{maintable_apibank}
\end{table}

\subsection{Main Results} We present the results of our proposed method and other baselines in Table \ref{maintable} and Table \ref{maintable_apibank}. The results show that our proposed two strategies, TDCL and RDS, outperform all the other dynamic sampling methods. Besides, the DSCL method, which is combined with these two strategies, achieves the best performance. The experimental results prove the effectiveness of our method, which brings a significant improvement over the original method.

Additionally, we find that it's hard to improve the performance of the model on the tool learning task by directly using the existing dynamic sampling methods such as DAPO. These methods are mainly designed for binary reward problems. Due to the gap in tasks, the tool learning does not require an intermediate problem-solving process, and the response length is relatively short. Thus, some of their techniques, such as the Soft Overlong Punishment of DAPO, are inferior in tool learning. Besides, another reason is that dynamic sampling on tool learning requires warmup first and cannot start sampling directly, which is analyzed in detail in the Table \ref{curritable}. 
This result highlights the need for a specially designed dynamic sampling method to address this task.

Moreover, experimental results show that our method improves the performance of each sub-task except the Relevance Detection, especially on the Multi Turn sub-task. Compared with other sub-tasks, the Multi Turn is more difficult. Our method gradually filters out data that is more valuable to the model during training, thereby enhancing the model to focus on difficult examples during training and improving the model's performance.

\begin{figure}[t]  
\centering
\includegraphics[width=0.45\textwidth]{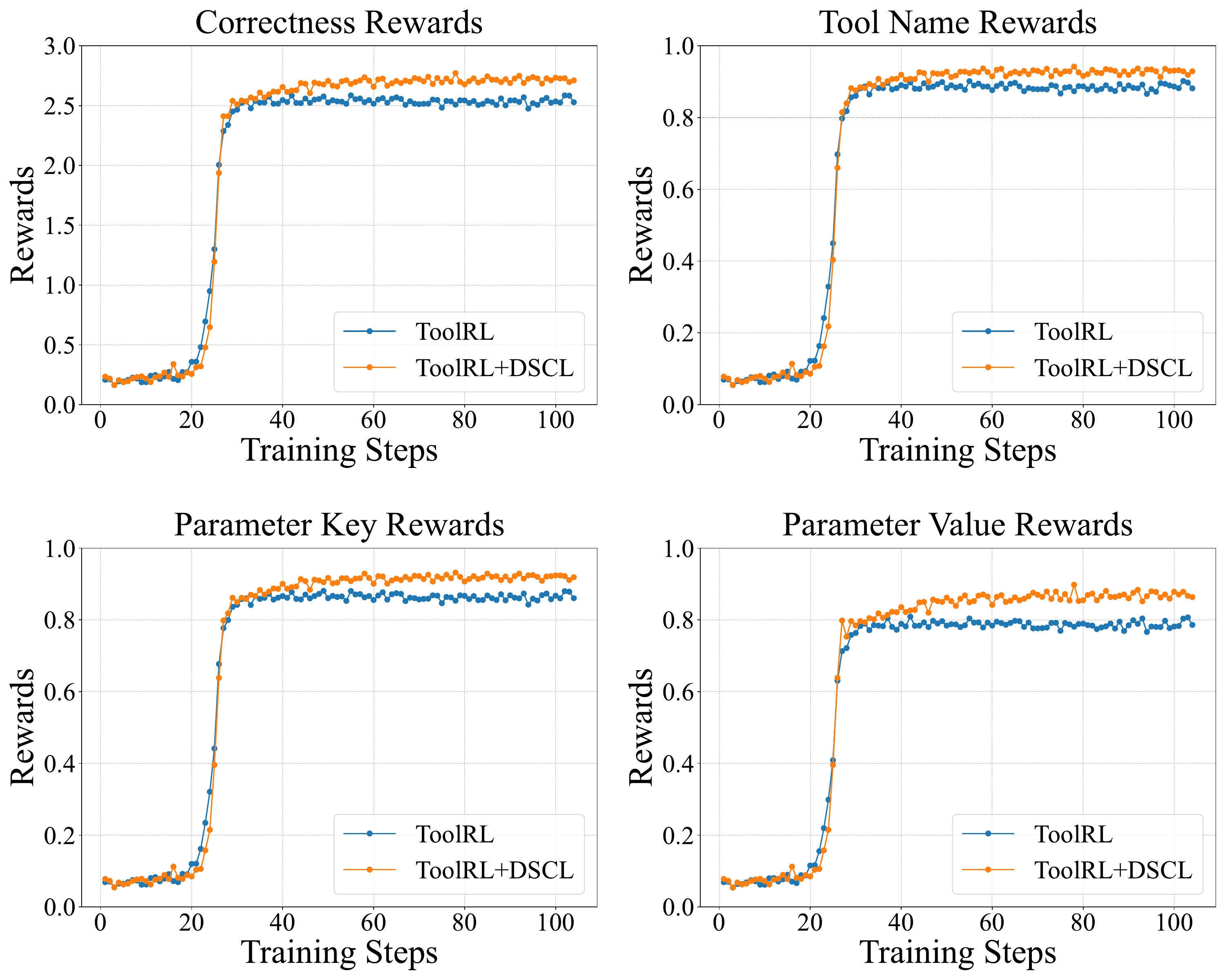} 
\caption{Training rewards of each sub-task. We have mapped the values of each reward to the interval $[0, 1]$ for each sub-tasks. The correctness rewards represent the total reward of the three sub-tasks.} 
\label{fig_curriculum}
\end{figure}

\subsection{Training Analysis} 
In this section, we will analyze the training process of our method in detail. 
As shown in Figure \ref{fig_curriculum},  we present the reward curves from the baseline ToolRL training compared to the one integrated with our DSCL method. It can be seen that after the introduction of our method, each sub-task can still show an upward trend after the training stabilizes, which demonstrates that our method can continuously provide valuable data to the model through the dynamic sampling strategy. 
Besides, on the more difficult sub-task, the parameter value task, our method shows a greater improvement compared with the original method, which is due to the fact that our curriculum learning method makes the model pay more attention to this sub-task in the mid-to-late training stage. At the same time, cooperating with the dynamic sampling method to select the corresponding data for training for this more difficult sub-task also helps. This result proves the effectiveness of our method.



\subsection{Data Analysis}
In this section, we conduct a more in-depth analysis of data distribution on the tool learning task. 

\begin{figure}[t]  
\centering
\includegraphics[width=0.5\textwidth]{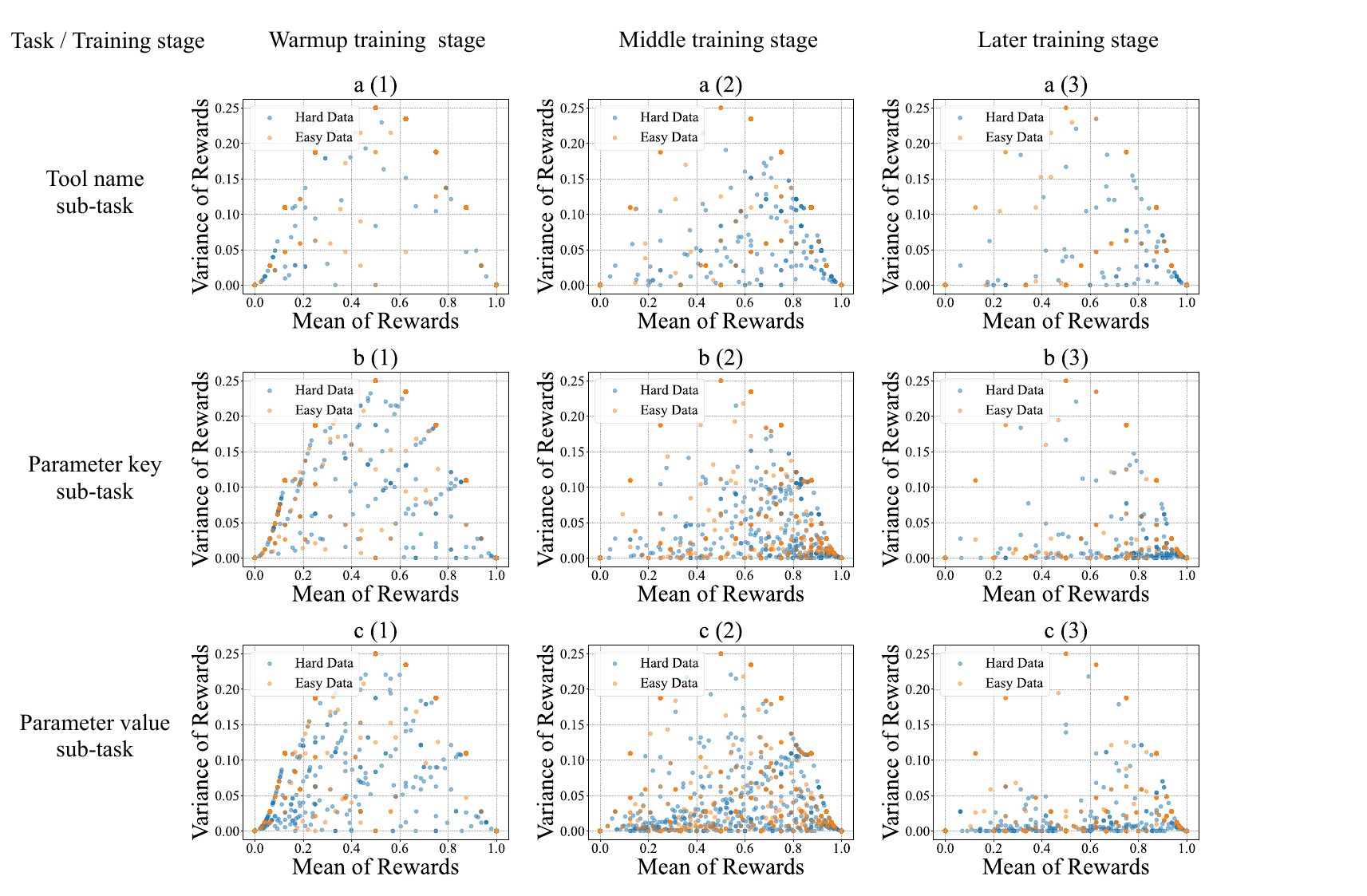} 
\caption{Mean-Variance Scatter Plots of data of different difficulty levels training rewards across different stages of ToolRL training. We have mapped the values of each reward to the interval $[0, 1]$.} 
\label{fig_base}
\end{figure}

We first analyze the model trained with all the data. Figure \ref{fig_base} illustrates the distribution of means and variances of training data with varying difficulty across different training stages and sub-tasks. Specifically, we classify the difficulty of the data based on the number of tools, the number of tool parameters, and the number of dialogue turns, with higher values of these metrics corresponding to increased data complexity. We quantified the counts of easy data and hard data instances as 2293 and 1627, respectively. It can be found that, compared to easy data, the distribution of hard data across each sub-plot exhibits characteristics of lower means and higher variances. Furthermore, this phenomenon becomes more pronounced with the progression of training.
Therefore, these samples should be given greater emphasis in the later stages of training; otherwise, the model will struggle to overcome its performance bottleneck.

\begin{figure}[t]  
\centering
\includegraphics[width=0.5\textwidth]{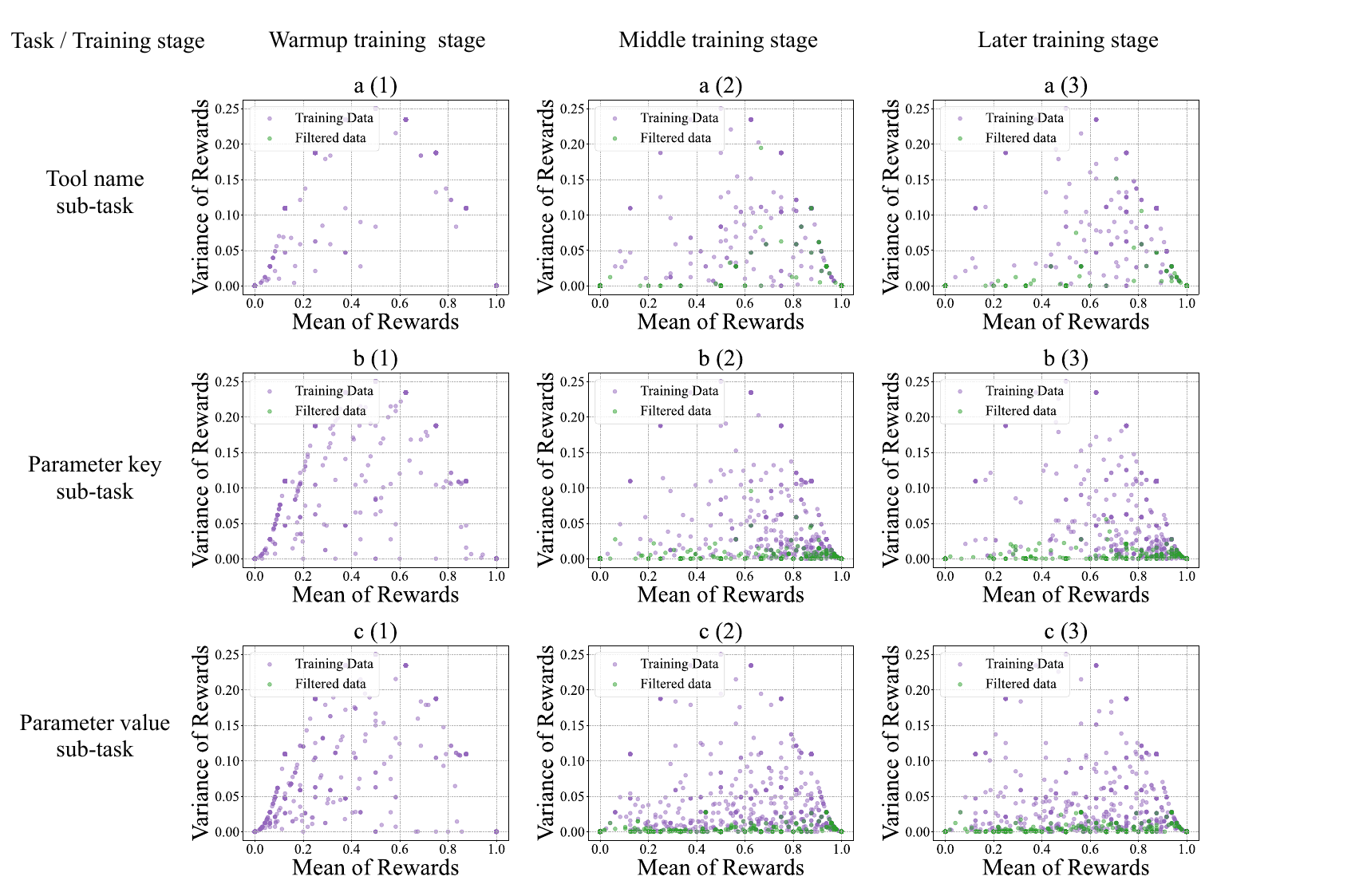} 
\caption{Mean-Variance Scatter Plots of training data (purple points) and filtered data (green points) rewards across different stages of DSCL training. We have mapped the values of each reward to the interval $[0, 1]$.} 
\label{fig_our}
\end{figure}

Based on this analysis, we recorded samples that were selected for training versus those that were excluded by our DSCL method for comparison. As illustrated in Figure~\ref{fig_our}, the majority of the data filtered out by DSCL shows a significant overlap with the simple samples presented in Figure~\ref{fig_base}. Furthermore, excluding the initial warmup phase where no sampling occurs, we tallied the number of samples used in training versus those discarded in the subsequent two stages. The counts were $(2243, 1397)$ and $(1950, 1690)$, respectively, indicating a gradual decrease in the quantity of data deemed valuable. These comparisons confirm that the DSCL method successfully and continuously focuses on valuable and challenging samples throughout training, thereby guiding the model towards a more optimal set of parameters.

\subsection{Training Tips on RDS}
\label{tips}
Tool learning task requires strict response formatting to extract tool information in a structured manner. As shown in Table \ref{curritable}, directly applying dynamic sampling leads to low reward scores due to formatting errors, because this causes the model to lose access to most training data, resulting in performance comparable to the baseline Qwen2.5-7B-Instruct model.
Therefore, we incorporate curriculum learning into our dynamic sampling approach. Specifically, we disable dynamic sampling during the initial warmup stage of training. The RDS method is activated only when the training reward stabilizes, as determined through dynamic monitoring. This curriculum learning strategy enables the model to effectively leverage the benefits of dynamic sampling.

\begin{table}[t]
\centering
\small
\setlength{\tabcolsep}{0.8mm}
\begin{tabular}{lcccc}
\toprule
Model                  & \makecell{Overal\\ Acc} & \makecell{Non-Live AST\\ Acc} & \makecell{Live\\ Acc} & \makecell{Multi Turn\\ Acc}  \\\midrule
Qwen2.5-7B-Instruct    & 47.68  & 68.98    & 63.31     & 8.88           \\
ToolRL   & 56.96  & 85.21  & 73.39  & 13.25    \\
\ \ \ \ +RDS w/o CL & 47.34 & 68.46 & 64.02 & 7.75   \\
\ \ \ \ +RDS & 60.48  & 87.27   & 76.63   & 18.25    \\
\bottomrule
\end{tabular}
\caption{
Ablation Study on the Impact of Curriculum Learning (CL) in RDS. 
Due to space limitations, we only show the most representative metrics.
}
\label{curritable}
\end{table}

\section{Conclusion}

In this work, we design a Dynamic Sampling method with Curriculum Learning (DSCL) for RL-based tool learning task. 
Our approach addresses the gaps in existing dynamic sampling methods for tool learning task.
The DSCL method contains two strategies, reward-based dynamic sampling (RDS) and task-based dynamic curriculum learning (TDCL).
The RDS strategy mainly focuses on the dynamic sampling the training data based on the rewards of all samples generated with the GRPO method. RDS selects samples based on the mean and variance of each data's samples and the variance of rewards for each data across different epochs. Meanwhile, we have incorporated a curriculum learning approach to fully leverage the performance of RDS in tool learning task.
As for the TDCL strategy, based on the characteristics of the tool learning task, we have categorized its sub-tasks into three levels of difficulty and dependency. We adopt a curriculum learning approach, progressively training on tasks of varying difficulty levels in stages.
Our training methodology thoroughly considers the characteristics of the tool learning task, which has a significantly improvement of the model's performance.
Meanwhile, the detailed analyses and ablation studies have validated our motivation and demonstrated the efficacy of our approach.

\bibliography{aaai2026}

\appendix
\section{Appendix}

\subsection{Experiment Settings}

\subsubsection{hyperparameter}
For our training process, we mainly follow the setting of ToolRL, all the modifications and parameters of our method are shown in the Table \ref{tab:parameter}:

\begin{table}[htbp]
\centering
\begin{tabular}{lc}
\toprule
\textbf{Hyperparameter} & \textbf{Value} \\
\midrule
\multicolumn{1}{l}{\emph{\textbf{Modifications:}}}     &  \\
Number of Rollouts & 8 \\ \midrule
\multicolumn{1}{l}{\emph{\textbf{Our Method:}}}     &  \\
$t_{mean}$ & 0.5 \\
$t_{var}$ & 0.1 \\
\bottomrule
\end{tabular}
\caption{Statistics of the API-Bank Evaluation Set.}
\label{tab:parameter}
\end{table}

All the models are trained with GRPO method using the veRL \cite{verl} framework and conducted on 8 H20 GPUs with 96GB of RAM for 15 epochs. 

\subsubsection{Dataset}
In this section, we give the detailed information of our training dataset in Table \ref{tab:rl_dataset_summary} and two evaluation dataset, BFCL in Table \ref{tab:bfcl_benchmark_enhanced_revised} and API-Bank in Table \ref{tab:api_bank_stats}.

\begin{table}[t]
\centering
\begin{tabular}{lc}
\toprule
\textbf{Statistic} & \textbf{Count} \\
\midrule
Domains & 8 \\
APIs & 73 \\
Dialogues & 314 \\
Turns & 914 \\
\quad Single-Call Turns & 363 \\
\quad Multi-Call Turns & 122 \\
\midrule
Call Tasks & 214 \\
Retrieve+Call Tasks & 50 \\
Plan+Retrieve+Call Tasks & 50 \\
\midrule
Avg. Turns per Dialogue & 2.91 \\
\bottomrule
\end{tabular}
\caption{Statistics of the API-Bank Evaluation Set.}
\label{tab:api_bank_stats}
\end{table}

\begin{table*}[t]
\centering
\begin{tabular}{llc}
\toprule
\textbf{Dataset} & \textbf{Key Contribution} & \textbf{Samples} \\
\midrule
ToolACE & \makecell[l]{\textbf{Diversity \& Complexity}: Covers single, parallel, dependent,\\ and non-tool-use scenarios to teach \textit{when} and \textit{how} to use tools.} & 2,000 \\
\addlinespace[0.5em]
Hammer (Masked) & \makecell[l]{\textbf{Semantic Reasoning}: Uses function masking to force the model\\ to understand API descriptions instead of relying on names.} & 1,000 \\
\addlinespace[0.5em]
xLAM & \makecell[l]{\textbf{Strategic Planning}: Features complex tasks requiring\\ orchestration of multiple tools and management of dependencies.} & 1,000 \\
\bottomrule
\end{tabular}
\caption{RL Training Dataset Composition and Rationale. We adopt the data strategy from ToolRL, combining three specialized datasets for a total of 4,000 training samples. Each dataset is chosen to foster a specific capability: ToolACE for foundational diversity, Hammer for semantic reasoning, and xLAM for advanced strategic planning.}
\label{tab:rl_dataset_summary}
\end{table*}

\begin{table*}[t]
\centering
\small
\setlength{\tabcolsep}{1mm}
\begin{tabular}{lclcl}
\toprule
\textbf{Main Category} & \textbf{Subtotal} & \textbf{Evaluation Dimension} & \textbf{Count} & \textbf{Evaluation Purpose} \\
\midrule

\multirow{2}{*}{\textbf{Non-Live (Single-Turn)}} & \multirow{2}{*}{\textbf{1,390}} & Non-Live AST & 1,150 & To test for \textbf{syntactic correctness}. \\
\cmidrule(lr){3-5}
& & Irrelevance Detection (Static) & 240 & To test for \textbf{hallucination avoidance}. \\
\midrule

\multirow{3}{*}{\textbf{Live (Single-Turn)}} & \multirow{3}{*}{\textbf{2,251}} & Live AST & 1,351 & To test for \textbf{executable correctness}. \\
\cmidrule(lr){3-5}
& & Relevance Detection (Live) & 18 & To test for \textbf{correct tool selection}. \\
\cmidrule(lr){3-5}
& & Irrelevance Detection (Live) & 882 & To test for \textbf{hallucination avoidance in a live context}. \\
\midrule

\textbf{Multi-Turn} & \textbf{800} & Multi-Turn Dialogue & 800 & To test for \textbf{stateful and sequential reasoning}. \\

\bottomrule
\end{tabular}
\caption{A Visually Enhanced Breakdown of the BFCL Benchmark (Revised Layout)}
\label{tab:bfcl_benchmark_enhanced_revised}
\end{table*}

\subsection{Tool Complexity and Task Difficulty Analysis}
We analyze how task complexity factors affect training performance across epochs. Specifically, we examine two complexity dimensions: (1) the number of available tool APIs and (2) the number of tool parameters. Our analysis covers four task categories: the overall task, tool name selection, parameter key extraction, and parameter value completion. 

\begin{figure}[t] 
\centering
\includegraphics[width=0.5\textwidth]{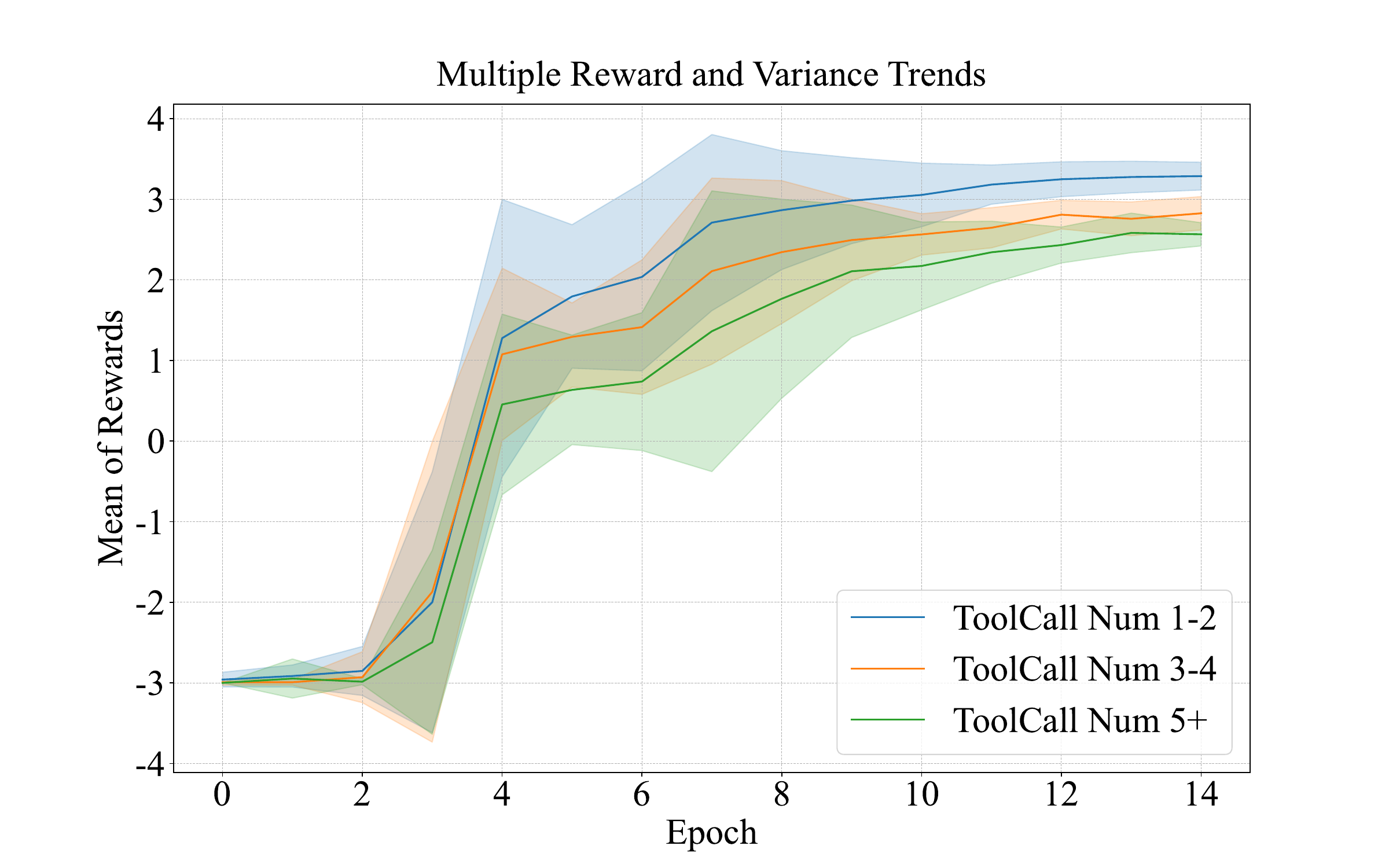} 
\caption{The relationship between training epochs and overall mean reward (the lines) and their corresponding variances (the shaded area) in ToolRL, categorized by the number of tool APIs. } 
\label{fig:tool_num}
\end{figure}

For mean reward, both complexity factors show consistent trends across all task categories. As the number of tool APIs or tool parameters increases, mean rewards consistently decrease during training 
(Figure \ref{fig:tool_num}).
This suggests that increased tool complexity directly reduces task performance.

\begin{figure}[t] 
\centering
\includegraphics[width=0.5\textwidth]{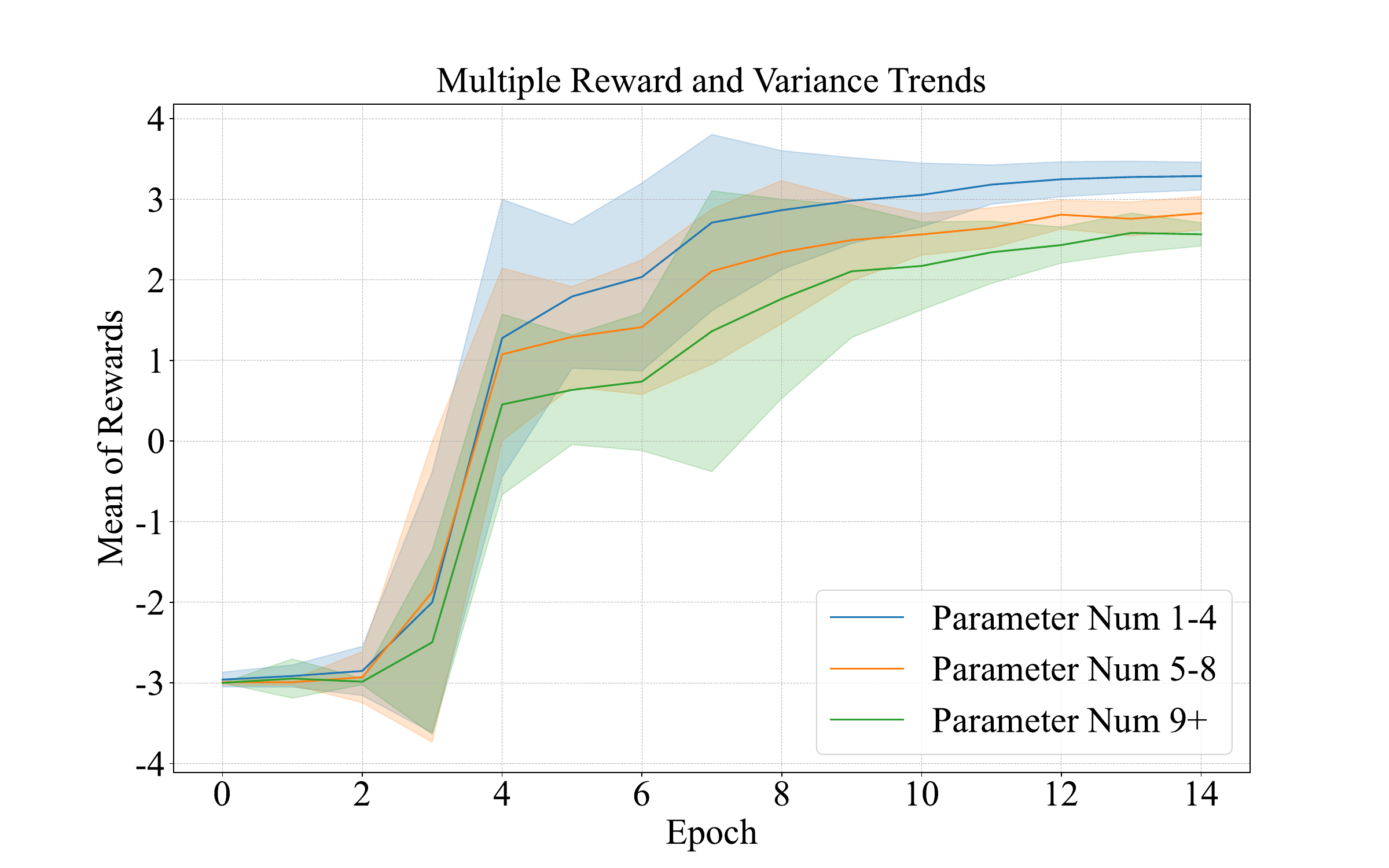} 
\caption{The relationship between training epochs and overall mean reward (the lines) and their corresponding variances (the shaded area) in ToolRL, categorized by the number of tools' parameters.} 
\label{fig:param_num}
\end{figure}

For the variance of reward, the opposite trend emerges. As the number of tool APIs or tool parameters increases, the reward variance increases consistently across all categories 
(Figure \ref{fig:param_num}). 
This indicates that complex tool environments introduce greater performance instability and learning difficulty.

Together, these findings demonstrate that task difficulty increases proportionally with both the size of the available tool set and the complexity of individual tool parameters.
 

\subsection{Prompts}
We follow the designation of prompt of ToolRL \cite{qian2025toolrl}. The specific system prompt design is shown in the Table \ref{systemprompt} and user prompt is shown in Table \ref{userprompt}.

\begin{table*}[t]
\begin{tabular}{l}
\toprule
You are a helpful multi-turn dialogue assistant capable of leveraging tool calls to solve user tasks and provide structured\\ chat responses.\\
\\
**Available Tools**\\
In your response, you can use the following tools:\\
1. Name: getGastroenterologyReport\\
Description: Retrieve gastroenterology report for a patient\\
Parameters: \{"patient\_id": \{"description": "The unique identifier of the patient", "type": "string", "default": ""\}, ...\}\\
2. ...\\
\\
**Steps for Each Turn**\\
1. **Think:** Recall relevant context and analyze the current user goal.\\
2. **Decide on Tool Usage:** If a tool is needed, specify the tool and its parameters.\\
3. **Respond Appropriately:** If a response is needed, generate one while maintaining consistency across user queries.\\
\\
**Output Format**\\
$<$think$>$ Your thoughts and reasoning $<$/think$>$\\
$<$tool\_call$>$\\
\{"name": "Tool name", "parameters": \{"Parameter name": "Parameter content", "... ...": "... ..."\}\}\\
\{"name": "... ...", "parameters": \{"... ...": "... ...", "... ...": "... ..."\}\}\\
...\\
$<$/tool\_call$>$\\
$<$response$>$ AI's final response $<$/response$>$\\
\\
**Important Notes**\\
1. You must always include the `$<$think$>$` field to outline your reasoning. Provide at least one\\ of `$<$tool\_call$>$` or `$<$response$>$`. Decide whether to use `$<$tool\_call$>$` (possibly multiple times),\\ `$<$response$>$`, or both.\\
2. You can invoke multiple tool calls simultaneously in the `$<$tool\_call$>$` fields. Each tool call\\ should be a JSON object with a "name" field and an "parameters" field containing a dictionary of\\ parameters. If no parameters are needed, leave the "parameters" field an empty dictionary.\\
3. Refer to the previous dialogue records in the history, including the user's queries, previous\\ `$<$tool\_call$>$`, `$<$response$>$`, and any tool feedback noted as `$<$obs$>$` (if exists).\\
\bottomrule
\end{tabular}
\caption{
System prompt
}
\label{systemprompt}
\end{table*}

\begin{table*}[t]
\centering
\begin{tabular}{l}
\toprule
**Dialogue History**\\
$<$user$>$ \{\{ Initial User Input \}\} $<$/user$>$\\
\\
$<$think$>$ Round 1 Model Thought $<$/think$>$\\
\{\{ Round 1 model output $<$tool\_call$>$ or $<$response$>$ \}\}\\
$<$obs$>$ Round 1 Observation $<$/obs$>$\\
... ...\\
\\
$<$user$>$ \{\{ User Input \}\} $<$/user$>$\\
... ...\\
\bottomrule
\end{tabular}
\caption{
User prompt
}
\label{userprompt}
\end{table*}

\FloatBarrier 
\end{document}